\documentclass[runningheads]{llncs}
\usepackage{amsmath}
\usepackage{amsfonts}
\usepackage{amssymb}
\usepackage{mathtools}
\usepackage{caption} 
\usepackage{multirow}
\usepackage{graphicx}
\graphicspath{{images/}}
\usepackage{wrapfig}
\usepackage{float}
\usepackage{tikz}
\usetikzlibrary{shapes.geometric} % for selector node
\usetikzlibrary{positioning, intersections}
\usetikzlibrary{calc}
\usetikzlibrary{fit}
\usepackage{amsmath}

\tikzset{mainstyle/.style={fill=white, draw=black, shape=rectangle, align=center}}

\tikzset{dstyle/.style={mainstyle, minimum size=4mm, inner sep=0pt, text width=5mm}}

\tikzset{sstyle/.style={mainstyle, minimum size=5mm, inner sep=0pt, text width=5mm}}

\tikzset{ostyle/.style={fill=black, draw=black, shape=rectangle, minimum size=0.2cm, inner sep=0pt, text width=2mm}}

\tikzset{fstyle/.style={solid, shape=circle, minimum size=0.6cm, scale=0.45, draw=black, outer sep=10pt, inner sep=1pt, line width=0.15mm, color=green, text=green}}

\tikzset{bstyle/.style={shape=circle, minimum size=0.6cm, scale=0.45, draw=black, outer sep=10pt, inner sep=1pt, line width=0.15mm, color=red, text=red}}

\tikzset{mstyle/.style={shape=circle, minimum size=0.6cm, scale=0.45, draw=black, outer sep=10pt, inner sep=1pt, line width=0.15mm}}

\tikzset{eqstyle/.style={mainstyle, minimum size=3mm, inner sep=0pt, outer sep=0pt, text width=3mm}}

\tikzstyle{observation}=[ostyle]
\tikzstyle{deterministic}=[dstyle]
\tikzstyle{stochastic}=[sstyle]
\tikzstyle{equality}=[eqstyle]
\tikzstyle{msg}=[mstyle]
\tikzstyle{fmsg}=[fstyle]
\tikzstyle{bmsg}=[bstyle]

\tikzstyle{filter}=[mainstyle, minimum width=1cm, minimum height=0.5cm]
\tikzstyle{selector}=[fill=white, draw=black, shape=trapezium, rotate=180, minimum width=1cm, minimum height=0.5cm]
\usepackage{hyperref}

\newcommand*\circled[1]{\tikz[baseline=(char.base)]{
            \node[shape=circle,draw,minimum size=4mm,inner sep=0pt] (char) {#1};}}
\newcommand*\squared[1]{\tikz[baseline=(char.base)]{
            \node[shape=rectangle,draw,minimum size=3.5mm,inner sep=0pt] (char) {#1};}}

\def\u{{\mathbf u}}

\def\y{{\mathbf y}}
\def\z{{\mathbf z}}
\def\m{{\mathrm m}}

\def\c{{\mathrm c}}
\def\d{{\mathrm d}}

\def\a{{\mathrm a}}
\def\b{{\mathrm b}}

\begin{document}
\title{Online system identification in a Duffing oscillator by free energy minimisation}
\titlerunning{Online system identification by free energy minimisation}
% If the paper title is too long for the running head, you can set
% an abbreviated paper title here
%
\author{Wouter M. Kouw%\inst{1}%\orcidID{0000-0002-0547-4817} %\and
% % Second Author\inst{2,3}\orcidID{1111-2222-3333-4444} \and
% % Third Author\inst{3}\orcidID{2222--3333-4444-5555}
% }
% \author{Authors%\inst{1}%\orcidID{0000-0002-0547-4817} %\and
% Second Author\inst{2,3}\orcidID{1111-2222-3333-4444} \and
% Third Author\inst{3}\orcidID{2222--3333-4444-5555}
}
\authorrunning{W.M. Kouw}
% First names are abbreviated in the running head.
% If there are more than two authors, 'et al.' is used.
%
\institute{Bayesian Intelligent Autonomous Systems lab\\ TU Eindhoven, Eindhoven, 5612AP Netherlands\\
\email{w.m.kouw@tue.nl}}
% \institute{Institute}{}
%
\maketitle              % typeset the header of the contribution
\begin{abstract}
Online system identification is the estimation of parameters of a dynamical system, such as mass or friction coefficients, for each measurement of the input and output signals. Here, the nonlinear stochastic differential equation of a Duffing oscillator is cast to a generative model and dynamical parameters are inferred using variational message passing on a factor graph of the model. The approach is validated with an experiment on data from an electronic implementation of a Duffing oscillator. The proposed inference procedure performs as well as offline prediction error minimisation in a state-of-the-art nonlinear model.
\end{abstract}
\section{Introduction}
% Natural agents start their lives without perfect control over their movement apparatus.
% Many species of mammals 
Natural agents are believed to develop an internal model of their motor system by generating actions in muscles and observing limb movements \cite{de2015baby}.
% \cite{saegusa2009active,nguyen2011model}. 
% Over time, the agent figures out what signal to send to assume a pose or to manipulate an object. 
% It learns a model that predicts future states from controls \cite{nguyen2011model}. 
It has been suggested that forming this internal model is analogous to a form of online system identification \cite{tin2005internal}. System identification, i.e. estimating dynamical parameters from observed input and output signals, has a rich history in engineering. But there might still be much to gain from considering biologically-plausible procedures. Here, I explore online system identification using a leading theory of how brains process information: free energy minimisation \cite{friston2006free,buckley2017free}.

To test free energy minimisation for use in engineering applications, I consider a specific benchmark\footnote{\url{http://nonlinearbenchmark.org/}} problem called a Duffing oscillator. Duffing oscillators are relatively well-behaved nonlinear differential equations, making them excellent toy problems for methodological research. Its differential equation is cast to a generative model, with a corresponding factor graph. The factor graph admits a recursive parameter estimation procedure through message passing
% : at each point in time, messages from the previous time step act as priors to be combined with messages from likelihood nodes to form posteriors
\cite{loeliger2007factor,korl2005factor}. Specifically, variational message passing minimises free energy \cite{dauwels2007variational,van2018forneylab,parr2019neuronal}. Here, I infer the parameters of a Duffing oscillator using online variational message passing. Experiments show that it performs as well as a nonlinear ARX model with parameters trained offline using prediction error minimisation \cite{aguirre2009modeling}.

\section{System}
\begin{figure}[htb]
    \centering
    \includegraphics[height=90px]{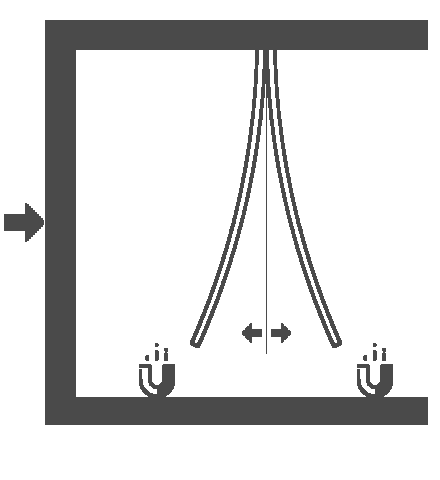} \quad
    \includegraphics[height=90px]{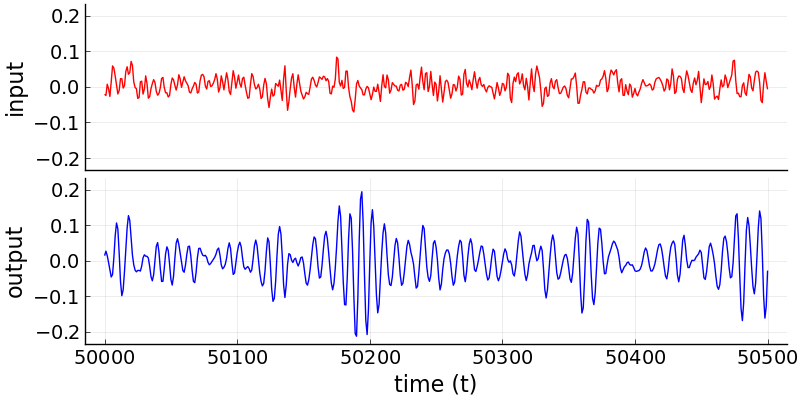}
    \caption{(Left) Example of a physical implementation of a Duffing oscillator. (Right) Example of input and output signals.}
    \label{fig:Duffing}
\end{figure}

Consider a rigid frame with two prongs facing rightwards (see Figure \ref{fig:Duffing} left). A steel beam is attached to the top prong. If the frame is driven by a periodic forcing term, the beam will displace horizontally as a driven damped harmonic oscillator. Two magnets are attached to the bottom prong, with the steel beam suspended in between. These act as a nonlinear feedback term on the beam's position, attracting or repelling it as it gets closer \cite{moon1979magnetoelastic}.

Let $y(t)$ be the observed displacement, $x(t)$ the true displacement, and $u(t)$ the observed driving force. The position of the beam is described as follows \cite{wigren2013three}:
\begin{subequations} \label{eq:ssm}
\begin{align}
    \m \frac{d^2 x(t)}{dt^2} + \c \frac{d x(t)}{dt} + \a x(t) + \b x^3(t) =&\ u(t) + w(t) \\ 
    y(t) =&\ x(t) + v(t) \, \, ,
\end{align}
\end{subequations}
where $\m$ is mass, $\c$ is damping, $\a$ the linear and $\b$ the nonlinear spring stiffness coefficient. Both the state transition as well as the observation likelihood contain noise terms, which are assumed to be Gaussian distributed: $w(t) \sim \mathcal{N}(0, \tau^{-1})$ (process noise) and $v(t) \sim \mathcal{N}(0, \xi^{-1})$ (measurement noise).  
The challenge is to estimate $\m$, $\c$, $\a$, $\b$, $\tau$ and $\xi$ such that the output of the system can be predicted as accurately as possible.

\section{Identification}
First, I discretise the state transition of Equation \ref{eq:ssm} using a central difference for the second derivative and a forward difference for the first derivative.
% \begin{align}
% \m (x_{t+1} - 2x_{t} + x_{t-1}) + \c (x_{t+1} - x_{t}) + \a x_t + \b x_t^3 =&\ u_t + w_t
% \end{align}
% where . 
Re-arranging to form an expression in terms of $x_{t+1}$ yields:
\begin{align}
x_{t+1} = \frac{2\m + \c \delta - \a \delta^2}{\m + \c \delta} x_{t} + \frac{-\b \delta^2}{\m + \c \delta}x_t^3 + \frac{-\m}{\m + \c \delta} x_{t-1} + \frac{\delta^2}{\m + \c \delta} (u_t + w_t) \, ,
\end{align}
where $\delta$ is the sample time step. Secondly, to ease inference at a later stage, I perform the following variable substitutions:
\begin{align}
\theta_1 \! = \! \frac{2\m \! + \! \c \delta \! - \! \a \delta^2}{\m \! + \! \c \delta} , \
\theta_2 \! = \! \frac{-\b \delta^2}{\m \! + \! \c \delta} , \
\theta_3 \! = \! \frac{-\m}{\m \! + \! \c \delta} , \
\eta \! = \! \frac{\delta^2}{\m \! + \! \c \delta} , \
\gamma \! = \! \frac{\tau (\m \! + \! \c \delta)^2}{\delta^4} ,
\end{align}
where the square in the numerator for $\gamma$ stems from absorbing the coefficient into the noise term ($\mathbb{V}[\eta w_t] = \eta^2\mathbb{V}[w_t]$). Note that the mapping between $\phi = (\m, \c, \a, \b, \tau)$ and $\psi = (\theta_1, \theta_2, \theta_3, \eta, \gamma)$ can be inverted to recover point estimates:
\begin{align} 
\m = \frac{-\theta_3 \delta^2}{\eta} \, , \ \
\c = \frac{(1+\theta_3)\delta}{\eta} \, , \ \
\a = \frac{1-\theta_1 - \theta_3}{\eta} \, , \ \
\b = \frac{-\theta_2}{\eta} \, , \ \
\tau = \gamma \eta^2 \, .
\end{align}
Thirdly, the state transition can be cast to a multivariate first-order form:
\begin{align} 
\underbrace{\begin{bmatrix} x_{t+1} \\ x_{t} \end{bmatrix}}_{z_t} = \underbrace{\begin{bmatrix} 0 \ & \ 0 \\ 1 \ & \ 0 \end{bmatrix}}_{S} \underbrace{\begin{bmatrix} x_{t} \\ x_{t-1} \end{bmatrix}}_{z_{t-1}} + \underbrace{\begin{bmatrix} 1 \\ 0 \end{bmatrix}}_{s} g(\theta, z_{t-1}) + \begin{bmatrix} 1 \\ 0 \end{bmatrix} \eta u_t + \begin{bmatrix} 1 \\ 0 \end{bmatrix} \tilde{w}_t \, ,
\end{align}
where $g(\theta, z_{t-1}) = \theta_1 x_t + \theta_2 x_t^3 + \theta_3 x_{t-1}$ and $\tilde{w}_t \sim \mathcal{N}(0, \gamma^{-1})$. The system is now a nonlinear autoregressive process.
% \begin{align}
%     z_t = f(z_{t-1}, \theta, \eta, u_t) + \tilde{w}_t
% \end{align}
% where $f(z_{t-1}, \theta, \eta, u_t) = Sz_{t-1} + e g(z_{t-1}, \theta) + e\eta u_t$. 
% Note that we need a two-dimensional state prior now (reminiscent of adding an initial condition on the velocity).
Lastly, integrating out $\tilde{w}_t$ and $v_t$ produces a Gaussian state transition and a Gaussian likelihood, respectively:
\begin{subequations}
\begin{align}\label{eq:nlarx}
z_t \sim&\ \mathcal{N}(f(\theta, z_{t-1}, \eta, u_t), V)  \\
y_t \sim&\ \mathcal{N}(s^{\top} z_t, \xi^{-1}) \, ,
\end{align}
\end{subequations}
where $f(\theta, z_{t-1}, \eta, u_t) = Sz_{t-1} + s g(\theta, z_{t-1}) + s \eta u_t$ and $V = \begin{bmatrix} \gamma^{-1}\ & 0\ ; 0 & \epsilon \end{bmatrix}$.
% and $W = V^{-1} = \begin{bmatrix} \gamma \ & \ 0 \\ 0 \ & \ \epsilon^{-1} \end{bmatrix}$. 
The number $\epsilon$ represents a small noise injection to stabilise inference \cite{dauwels2009expectation}.

To complete the generative model description, priors must be defined. Mass $\m$ and process precision $\tau$ are known to be strictly positive parameters, while the damping and stiffness coefficients can be both positive and negative. By examining the variable substitutions, it can be seen that $\theta_1$, $\theta_2$, $\theta_3$ and $\eta$ can be both positive and negative, but $\gamma$ can only be positive. As such, the following parametric forms can be chosen for the priors:
\begin{align}
\theta \sim \mathcal{N}(m^{0}_{\theta}, V^{0}_{\theta}) \, , \quad
\eta \sim \mathcal{N}(m^{0}_{\eta}, v^{0}_{\eta}) \, , \quad
\gamma \sim \Gamma(a^{0}_\gamma, b^{0}_\gamma) \, , \quad 
\xi \sim \Gamma(a^{0}_\xi, b^{0}_\xi) \, .
\end{align}
% Modeling $\theta = (\theta_1, \theta_2, \theta_3)$ as a joint Gaussian allows for incorporating it into an Autoregressive node (see Section \ref{sec:fg}.
% Measurement precision $\xi$ is modeled by a Gamma distribution.

\subsection{Free energy minimisation}
Given the generative model, a free energy functional with a recognition model $q$ can be formed as follows:
\begin{align}
    -\log p(\y, \u) \leq \iint q(\psi,\z) \frac{q(\psi,\z)}{p(\y, \u, \z, \psi)}\ \d \z \d \psi \ = \mathcal{F}[q]
\end{align}
where $\z = (z_1, \dots, z_T)$, $\y = (y_1, \dots, y_T)$ and $\u = (u_1, \dots, u_T)$. I assume the states factor over time and that the parameters are largely independent:
\begin{align}
    q(\psi, \z) = q(\theta) q(\eta) q(\gamma) q(\xi) \prod_{t=1}^{T} q(z_t) \, .
\end{align}
All recognition densities are Gaussian distributed, except for $q(\gamma)$ and $q(\xi)$, which are Gamma distributed. In free energy minimisation, the parameters of the recognition distributions depend on each other and are iteratively updated.
% We will execute a recursive variational inference scheme through message passing on a factor graph.

\subsection{Factor graphs and message passing} \label{sec:fg}
\begin{figure}[htb]
    \centering
    \scalebox{1.4}{\begin{tikzpicture}

% nodes first layer
\node [style=filter, minimum width=2cm] (AR) {NLARX};
\node [above left = -2mm and -3mm of AR] (ARxtprev) {};
\node [below right = -1mm and -3mm of AR] (ARxt) {};
\node [above = 2mm of ARxt] (ARgx) {};
\node [above right = 2mm and -10mm of ARxt] (ARex) {};
\node [above right =-1mm and -10mm of ARxt] (ARex2) {};
\node [above right = 2mm and -19mm of ARxt] (ARtx) {};
\node [above = 2mm of ARxtprev] (ARa) {};

\node [style=equality, above right = 17mm and -3.25mm of AR] (gammax) {=};
\node [left of = gammax, node distance=8mm] (gammaxprev) {};
\node [right of = gammax, node distance=8mm] (gammaxnext) {};
\node [style=equality, above right = 11.5mm and -11mm of AR] (etax) {=};
\node [left of = etax, node distance=8mm] (etaxprev) {};
\node [right of = etax, node distance=8mm] (etaxnext) {};
\node [style=equality, above right = 6mm and -20mm of AR] (thetax) {=};
\node [left of = thetax, node distance=8mm] (thetaxprev) {};
\node [right of = thetax, node distance=8mm] (thetaxnext) {};
\node [left of = AR, node distance=24mm] (xtprev) {$\dots$};
\node [left of = thetax, node distance=15.5mm] (thetadotsprev) {$\dots$};
\node [left of = etax, node distance=24.5mm] (etadotsprev) {$\dots$};
\node [left of = gammax, node distance=32.5mm] (gammadotsprev) {$\dots$};
\node [right of = thetax, node distance=43.25mm] (thetadotsnext) {$\dots$};
\node [right of = etax, node distance=34.25mm] (etadotsnext) {$\dots$};
\node [right of = gammax, node distance=26.5mm] (gammadotsnext) {$\dots$};
\node [style=equality, right of = AR, node distance=20mm] (xt) {=};
\node [right of = xt, node distance=15mm] (xtnext) {$\dots$};

% messages
\draw[->] [>=stealth] (gammax) -- node[style=msg, left] {$4$} node[left, inner sep=0pt] {$\scriptstyle{\downarrow}$} node[style=msg, right] {$8$} node[right, inner sep=0pt] {$\scriptstyle{\uparrow}$} (ARgx);
\draw[->] [>=stealth] (etax) -- node[style=msg, left] {$3$} node[left, inner sep=0pt] {$\scriptstyle{\downarrow}$} node[style=msg, right] {$7$} node[right, inner sep=0pt] {$\scriptstyle{\uparrow}$} (ARex);
\draw[->] [>=stealth] (thetax) -- node[style=msg, left] {$2$} node[left, inner sep=0pt] {$\scriptstyle{\downarrow}$} node[style=msg, right] {$6$} node[right, inner sep=0pt] {$\scriptstyle{\uparrow}$} (ARtx);
\draw[->] [>=stealth] (xtprev) -- node[above] {$\scriptstyle{z_{t-1}}$} node[style=msg, below] {$\mathbf{1}$} node[below, inner sep=1pt] {$\scriptstyle{\rightarrow}$} (AR);
\draw[->] [>=stealth] (AR) -- node[style=msg, below] {$9$} node[below, inner sep=1pt] {$\scriptstyle{\rightarrow}$} node[style=msg, above] {$5$} node[above, inner sep=1pt] {$\scriptstyle{\leftarrow}$} (xt);
\draw[->] [>=stealth] (xt) -- node[above] {$\scriptstyle{z_{t}}$} node[style=msg, below] {$12$} node[below, inner sep=1pt] {$\scriptstyle{\rightarrow}$} (xtnext);
\draw[dashed] (gammaxprev) -- (gammax);
\draw[dashed] (gammax) -- node[above] {$\scriptstyle{\gamma}$} (gammaxnext);
\draw[dashed] (etaxprev) -- (etax);
\draw[dashed] (etax) -- node[above] {$\scriptstyle{\eta}$} (etaxnext);
\draw[dashed] (thetaxprev) -- (thetax);
\draw[dashed] (thetax) -- node[above] {$\scriptstyle{\theta}$} (thetaxnext);

% observation
\node [style=deterministic, below of=xt, node distance=7mm] (cT) {$\cdot$};
\node [right =4mm of cT, node distance=10mm] (e) {$s$};
\node [style=stochastic, below of = cT, node distance=7mm] (ynoise) {$\mathcal{N}$};
% \node [above left= 2mm and 4mm of ynoise, node distance=0mm] (ygamma) {$\scriptstyle{\xi}$};
% \node [left of = ygamma, node distance=8mm] (ygammaprev) {};
% \node [right of = ygamma, node distance=8mm] (ygammanext) {};
\node [style=observation, below of=ynoise, node distance=6mm] (yt) {};
\node [below of=yt, node distance=3mm] (yobs) {$\scriptstyle{y_t}$};
\node [style=equality, above left = -2mm and 10mm of ynoise] (xi) {=};
\node [left of = xi, node distance=6mm] (xiprev) {};
\node [right of = xi, node distance=6mm] (xinext) {};
\node [left of = xi, node distance=30mm] (xidotsprev) {$\dots$};
\node [right of = xi, node distance=29mm] (xidotsnext) {$\dots$};
\node [left = 10mm of ynoise, node distance=0mm] (ygammacorner) {};

\draw [->] [>=stealth] (xt) -- (cT);
\draw [->] [dashed] (e) -- (cT);
\draw [->] [>=stealth] (cT) --  (ynoise);
% \draw [->] [>=stealth] (cT) -- node[style=msg, left] {$9$} node[left, inner sep=0pt] {$\scriptstyle{\downarrow}$} node[style=msg, right] {$11$} node[right, inner sep=0pt] {$\scriptstyle{\uparrow}$} (ynoise);
% \draw [-] [>=stealth] (xi) -- (ygammacorner);
% \draw [-] [>=stealth] (ynoise) -- node[style=msg, solid, below] {$10$} node[below, inner sep=1pt] {$\scriptstyle{\rightarrow}$} (ygammacorner);
\draw [->]  [>=stealth] (ynoise) -- node[left] {} (yt);
\draw[dashed] (xiprev) -- (xi);
\draw[dashed] (xi) -- node[above] {$\scriptstyle{\xi}$} (xinext);

\draw [->] (xi) |- node[style=msg, solid, below right=.5mm and 6mm] {$10$} node[below right=0mm and 6mm, inner sep=1pt] {$\scriptstyle{\rightarrow}$} node[style=msg, solid, above right=.5mm and 6mm] {$11$} node[above right=-0.1mm and 6mm, inner sep=1pt] {$\scriptstyle{\leftarrow}$} (ynoise);

% Control
\node [style=observation, below of=ARex, node distance=8mm] (ut) {};
\node [below of=ut, node distance=3mm] (uobs) {$\scriptstyle{u_t}$};
\draw [->]  [>=stealth] (ARex2) -- node[left] {} (ut);

% Priors
\node [style=stochastic, left of = AR, node distance=33.3mm] (priorx) {$\mathcal{N}$};
\node [style=stochastic, left of = thetax, node distance=25mm] (priortheta) {$\mathcal{N}$};
\node [style=stochastic, left of = etax, node distance=34mm] (prioreta) {$\mathcal{N}$};
\node [style=stochastic, left of = gammax, node distance=41.75mm] (priorgamma) {$\Gamma$};
\node [style=stochastic, left of = xi, node distance=39mm] (priorxi) {$\Gamma$};

\draw [->]  [>=stealth] (priorx) -- node[left] {} (xtprev);
\draw [->]  [>=stealth] (priortheta) -- node[left] {} (thetadotsprev);
\draw [->]  [>=stealth] (prioreta) -- node[left] {} (etadotsprev);
\draw [->]  [>=stealth] (priorgamma) -- node[left] {} (gammadotsprev);
\draw [->]  [>=stealth] (priorxi) -- node[left] {} (xidotsprev);

\end{tikzpicture}}
    \caption{Forney-style factor graph of the generative model of a Duffing oscillator. Nodes represent conditional distributions and edges represent variables. Nodes send messages to connected edges. When two messages on an edge collide, the marginal belief $q$ for the corresponding variable is updated. Each belief update reduces free energy. By iterating message passing, free energy is minimised.}
    \label{fig:ar-node}
\end{figure}
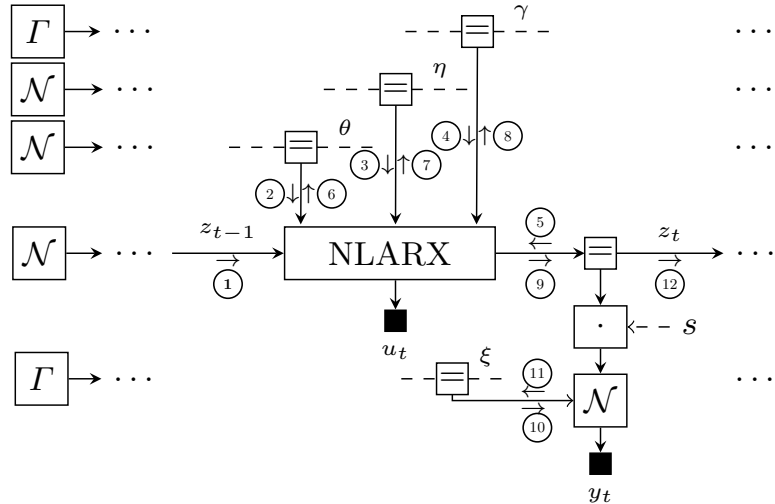

In online system identification, parameter estimates should be updated at each time-step. That puts time constraints on the inference procedure. Message passing is an ideal inference procedure due to its efficiency in factorised generative models \cite{korl2005factor}. Figure \ref{fig:ar-node} is a graphical representation of the generative model, with nodes for factors and edges for variables. Square nodes with Greek letters represent stochastic operations while $\squared{$\cdot$}$ and $\squared{$=$}$ represent deterministic operations. The node marked "NLARX" represents the state transition described in Equation \ref{eq:nlarx}.

The terminal nodes on the left represent the initial priors for the states and dynamical parameters. Inference starts when these nodes pass messages. The subgraph - separated by columns of dots - represents the structure of a single time step, recursively applied. Messages $\circled{1}, \circled{2}, \circled{3}, \circled{4}$ and $\circled{10}$ represent beliefs $q$ from previous time-steps. 
% These messages act as priors and combined with messages from nodes with observations, they form posteriors.
Message $\circled{5}$, arriving at the state transition node, originates from the likelihood node attached to observation $y_t$. Messages $\circled{6}, \circled{7}, \circled{8}$, $\circled{9}$ and $\circled{11}$ combine priors from previous time steps and likelihoods of observations, and are used to update beliefs $q$. 
% Message $\circled{11}$ updates the measurement precision belief $q(\xi)$. 
Message $\circled{12}$ is the current state belief and becomes message $\circled{1}$ in the next time step. 

The graph actually contains more messages, such as those sent by equality nodes. I have hidden them to avoid complicating the figure. Their form has been extensively described in the literature and can be looked up easily \cite{korl2005factor,loeliger2007factor}. Modern message passing toolboxes, such as Infer.NET and ForneyLab.jl, automatically incorporate them. However, the NLARX node is new. Its messages can be computed with\footnote{Derivations at \url{https://github.com/biaslab/IWAI2020-onlinesysid}}:
\begin{subequations} \label{eq:messages}
\begin{align} 
    \circled{6}&\ \overrightarrow{\nu}(\theta) = \exp \Big( \mathbb{E}_{q(z_t) q(z_{t-1}) q(\eta) q(\gamma)} \big[ \log \mathcal{N}(f(\theta, z_{t-1}, \eta, u_t), V) \big] \Big) \\
    \circled{7}&\ \overrightarrow{\nu}(\eta) = \exp \Big( \mathbb{E}_{q(z_t) q(z_{t-1}) q(\theta) q(\gamma)} \big[ \log \mathcal{N}(f(\theta, z_{t-1}, \eta, u_t), V) \big] \Big) \\
    \circled{8}&\ \overrightarrow{\nu}(\gamma) = \exp \Big( \mathbb{E}_{q(z_t) q(z_{t-1}) q(\theta) q(\eta)} \big[ \log \mathcal{N}(f(\theta, z_{t-1}, \eta, u_t), V) \big] \Big) \\
    \circled{9}&\ \overrightarrow{\nu}(z_t) = \exp \Big( \mathbb{E}_{q(z_{t-1}) q(\theta) q(\eta) q(\gamma)} \big[\log \mathcal{N}(f(\theta, z_{t-1}, \eta, u_t), V) \big] \Big) \, ,
\end{align}
\end{subequations}
where I use a first-order Taylor expansion to approximate the expected value of the nonlinear autoregressive function $g(\theta, z_{t-1})$.

Loeliger et al. (2007) have written an accessible introduction on message passing in factor graphs \cite{loeliger2007factor}. Variational message passing in autoregressive processes has been described in detail as well \cite{dauwels2007variational,podusenko2020online}.

\section{Experiment}
\begin{figure}[htb]
    \centering
    \includegraphics[width=.8\textwidth]{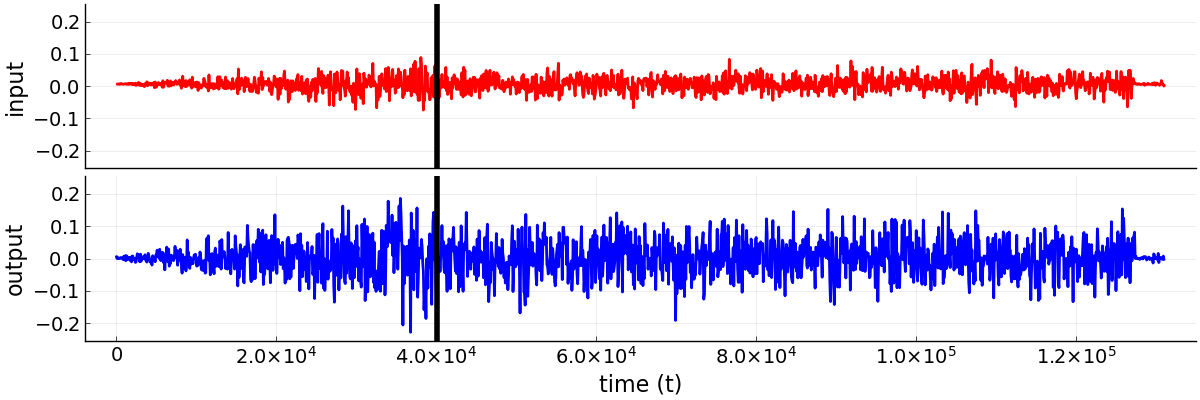}
    \caption{Silverbox data set, sampled at every $80$ time steps for visualisation. The black line splits it into validation data (left) and training data (right).}
    \label{fig:silverbox_data}
\end{figure}

The Duffing oscillator has been implemented in an electronic system called Silverbox \cite{wigren2013three}. It consists of $T$ = $131702$ samples, gathered with a sampling frequency of 610.35 Hz. Figure \ref{fig:silverbox_data} shows the time-series, plotted at every $80$ time steps. There are two regimes: the first $40 000$ samples are subject to a linearly increasing amplitude in the input (left of the black line in Figure \ref{fig:silverbox_data}) and the remaining samples are subject to a constant amplitude but contain only odd harmonics (right of the black line). The second regime is used as a training data set, where both input and output data were given and parameters needed to be inferred. The first regime is used as a validation data set, where the inferred parameters are fixed and the model needs to make predictions for the output signal.

I performed two experiments\footnote{Experiment notebooks at \url{https://github.com/biaslab/IWAI2020-onlinesysid}}: a 1-step ahead prediction error and a simulation error setting. I used ForneyLab.jl, with NLARX as a custom node, to run the message passing inference procedure \cite{cox2018forneylab}. I call the model above {\sc FEM-NLARX}, for Nonlinear Latent Autoregressive model with eXogenous input using Free Energy Minimisation. I implemented two baselines: the first is NLARX without the nonlinearity (i.e. the nonlinear spring coefficient $\b = 0$), dubbed {\sc FEM-LARX}. The second is a standard NARX model, implemented using MATLAB's System Identification Toolbox. I modelled the static nonlinearity with a sigmoid network of 4 units (in line with the 4 coefficients used by NLARX and LARX). Parameters were inferred offline using Prediction Error Minimisation. Hence, this baseline is called {\sc PEM-NARX}. 

I chose uninformative priors for the coefficients $\theta$ and $\eta$: Gaussians centred at $1$ with precisions of $0.1$. The authors of Silverbox indicate that the signal-to-noise ratio at measurement time was high \cite{wigren2013three}. I therefore chose informative priors for the noise parameters: $a_{\xi}^0 = 1e8$ and $a_{\gamma}^0 = 1e3$ (shape parameters) and $b_{\xi}^0 = 1e3$ and $b_{\gamma}^0 = 1e1$ (scale parameters).

\subsection{1-step ahead prediction error}
\begin{figure}[htb]
    \centering
    \includegraphics[width=.9\textwidth]{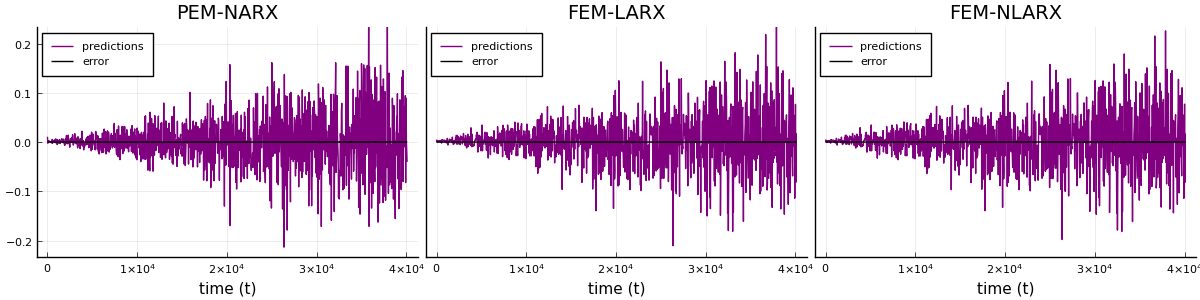}\\
    \includegraphics[width=.9\textwidth]{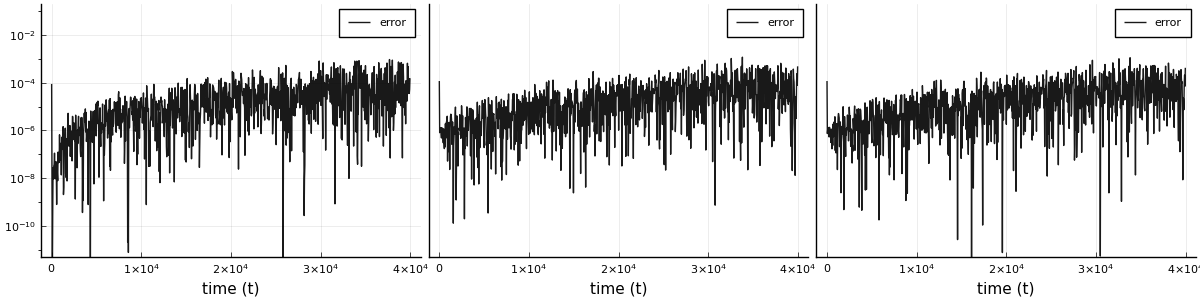}
    \caption{1-step ahead prediction errors. (Left) Offline NARX model with sigmoid net (PEM-NARX), (middle) online linear model (FEM-LARX) and (right) online nonlinear model (FEM-NLARX). (Top) Predictions (purple) and squared error (black). (Bottom) Squared prediction errors in log-scale.}
    \label{fig:prederror}
\end{figure}

At each time-step in the validation data, the models were given the previous output signal $y_{t-1}, y_{t-2}$ and the current input signal $u_t$ and had to infer the current output $y_t$. It is a relatively easy task, which is reflected in all three models' performance. The top row in Figure \ref{fig:prederror} shows the predictions of all three models in purple and their squared error with respect to the true output signal in black. The left column shows the offline NARX baseline (PEM-NARX), the middle column the linear online latent autoregressive baseline (FEM-LARX) and the right column the nonlinear online latent autoregressive model (FEM-NLARX). Note that the errors in the top row seem completely flat. The bottom row in the figure plots the errors on a log-scale. PEM-NARX has a mean squared error of $5.831$e-$5$, FEM-LARX one of $5.945$e-$5$ and FEM-NLARX one of $5.830$e-$5$.

\subsection{Simulation error}
\begin{figure}[htb]
    \centering
    \includegraphics[width=.9\textwidth]{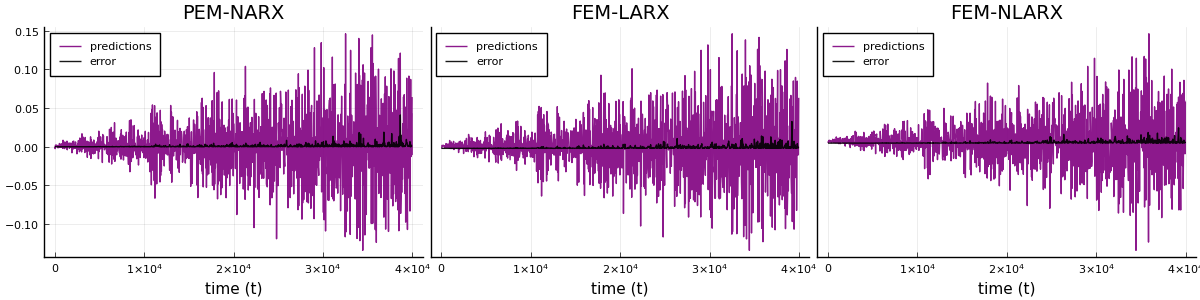}\\
    \includegraphics[width=.9\textwidth]{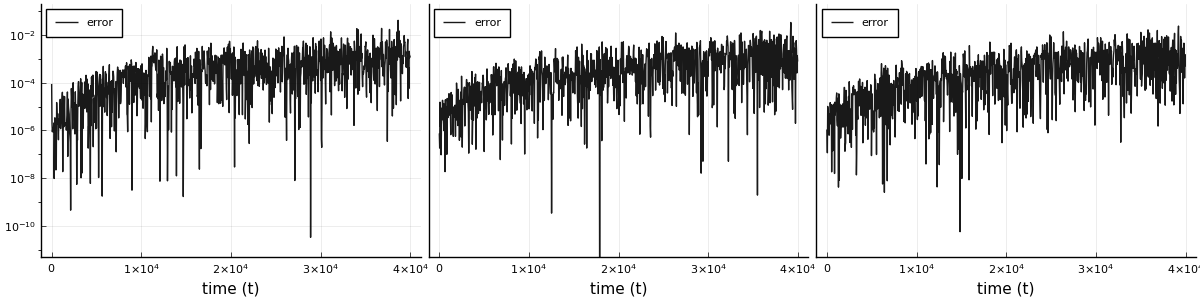}
    \caption{Simulation errors. (Left) Offline NARX model with sigmoid net (PEM-NARX), (middle) online linear model (FEM-LARX) and (right) online nonlinear model (FEM-NLARX). (Top) Predictions (purple) and squared error (black). (Bottom) Squared prediction errors in log-scale.}
    \label{fig:simerror}
\end{figure}

In this experiment, the models were not given the previous output signal, but had to use their predictions from the previous time-step. This is a much harder task, because errors will accumulate. The top row in Figure \ref{fig:simerror} again shows the predictions of all three models (purple) and their squared error (black). It can already be seen that the errors increase as the input signal's amplitude rises. The bottom row plots the errors on a log-scale. PEM-NARX has a mean squared error of $1.000$e-$3$, FEM-LARX one of $1.002$e-$3$ and FEM-NLARX one of $0.926$e-$3$.

\section{Discussion}
The experimental results seem to justify looking to nature for inspiration. Free energy minimisation, in the form of variational message passing, seems a generally applicable and well-performing inference technique. The difficulties mostly lie in deriving variational messages (i.e. Equations \ref{eq:messages}). 

Improvements in the proposed procedure could be made with a richer approximation of the nonlinear autoregressive function (e.g. unscented transform) \cite{sarkka2013bayesian}. Alternatively, a hierarchy of latent Gaussian filters or autoregressive processes could be used to obtain time-varying noise parameters or time-varying coefficients \cite{senoz2020bayesian,podusenko2020online}. Furthermore, instead of discretising such that an auto-regressive model is obtained, one could express the evolution of the states in generalised coordinates. Lastly, black-box models could be explored for further performance improvements.

% \subsection{Alternative modeling steps}
% The model shown above ends up being an autoregressive model because we started with discretisation before re-arranging the dynamics. Had we first mapped the higher-order differential equation into a first-order system and then discretised, then we would have ended up with a generalised filter. Future work will determine which is the more appropriate approach.

% \subsection{Optimal design by active inference}
A natural next step is for an active inference agent to determine the control signal regime (i.e. optimal design). Unfortunately, this is not straightforward: the current formulation relies on variational free energy which does not produce an epistemic term in the objective. The epistemic term is needed to encourage exploration; i.e. try sub-optimal inputs to reduce uncertainty. To arrive at an epistemic term, one would need to work with expected free energy \cite{parr2019generalised}. But it is unclear how expected free energy could be incorporated into factor graphs.
% An in-depth discussion with the active inference community could prove useful.

\subsection{Related work}
% System identification is a rich field with a long history \cite{ljung1999system}. In many engineering applications, data is collected once and offline estimation techniques, such as frequency response models, produce highly accurate fits \cite{pintelon2012system}. 
Online system identification procedures typically employ recursive least-squares or maximum likelihood inference, with nonlinearities modelled by basis expansions or neural networks \cite{paleologu2008robust,tangirala2018principles,engel2004kernel}. 
% But such procedures lack a probabilistic foundation. 
Online Bayesian identification procedures come in two flavours: sequential Monte Carlo samplers \cite{green2015bayesian,abdessalem2016identification} and online variational Bayes \cite{yoshimoto2003system,fujimoto2011system}. This work is novel in the use of variational message passing as an efficient implementation of online variational Bayes and its application to a nonlinear autoregressive model.

\section{Conclusion}
I have presented a free energy minimisation procedure for online system identification. Experimental results showed comparable performance to a state-of-the-art nonlinear model with parameters estimated offline. This indicates that the procedure performs well enough to be deployed in engineering applications. 

Future work should test variational message passing in more challenging nonlinear identification settings, such as a Wiener-Hammerstein benchmark \cite{schoukens2017three}. Furthermore, problems with time-varying dynamical parameters, such as a robotic arm picking up objects with mass, would be interesting for their connection to natural agents. 

\section{Acknowledgements}
The author thanks Magnus Koudahl, Albert Podusenko and Thijs van de Laar for insightful discussions and the reviewers for their constructive feedback.
\bibliographystyle{splncs04}
\bibliography{references}

\end{document}